\documentclass[11pt]{article}

\RequirePackage[OT1]{fontenc}
\RequirePackage{amsthm,amsmath}
\RequirePackage[numbers]{natbib}
\RequirePackage[colorlinks,citecolor=blue,urlcolor=blue]{hyperref}

\usepackage{amssymb,amsmath}
\usepackage{color}
\usepackage{graphicx}
\usepackage{tikz}

\def\eb{{\bf e}}

\def\G{{\cal G}}

\def\J{{\cal J}}

\def\N{{\cal N}}

\def\S{{\cal S}}

\def\R{{\mathbb R}}

\def\al{\alpha}
\def\d{\delta}

\def\e{\epsilon}
\def\g{\gamma}

\def\L{\Lambda}

\def\s{\sigma}
\def\SI{\Sigma}

\def\th{\theta}

\def\ap{\rightarrow}

\def\es{\emptyset}
\def\seq{\subseteq}

\def\fa{\; \forall}

\def\st{\mbox{ s.t. }}

\def\nm{\Vert}

\renewcommand{\and}{\mbox{$\wedge$}}

\newcommand{\bc}{\begin{center}}
\newcommand{\ec}{\end{center}}
\newcommand{\be}{\begin{equation}}
\newcommand{\ee}{\end{equation}}
\newcommand{\bd}{\begin{displaymath}}
\newcommand{\ed}{\end{displaymath}}
\newcommand{\ba}{\begin{array}}
\newcommand{\ea}{\end{array}}
\newcommand{\ben}{\begin{enumerate}}
\newcommand{\een}{\end{enumerate}}
\newcommand{\bit}{\begin{itemize}}
\newcommand{\eit}{\end{itemize}}
\newcommand{\beq}{\begin{eqnarray}}
\newcommand{\eeq}{\end{eqnarray}}
\newcommand{\btab}{\begin{tabular}}
\newcommand{\etab}{\end{tabular}}
\newcommand{\bfig}{\begin{figure}}
\newcommand{\efig}{\end{figure}}
\newcommand{\btp}{\begin{tikzpicture}}
\newcommand{\etp}{\end{tikzpicture}}

\newcommand{\halmos}{\hfill $\Box$}

\newcommand{\nmm}[1]{ \nm #1 \nm }
\newcommand{\nmeu}[1]{ \nm #1 \nm_2 }
\newcommand{\nmeusq}[1]{ \nm #1 \nm_2^2 }
\newcommand{\nmA}[1]{ \nm #1 \nm_A }
\newcommand{\nmP}[1]{ \nm #1 \nm_P }

\newcommand{\supp}{\mbox{supp}}
\def\xh{\hat{x}}

\def\xl{x_{\L}}

\def\xlo{x_{\L_0}}
\def\xloc{x_{\L_0^c}}
\def\hl{h_{\L}}
\def\hlc{h_{\L^c}}
\def\hlo{h_{\L_0}}
\def\hloc{h_{\L_0^c}}

\def\GkS{{\rm GkS}}

\def\nmsl1{\nm_{{\rm SL1}}}

\newtheorem{corollary}{Corollary}{\bf}{\it}
\newtheorem{definition}{Definition}{\bf}{\it}
{\bf}{\rm}
\newtheorem{lemma}{Lemma}{\bf}{\it}
\newtheorem{theorem}{Theorem}{\bf}{\it}
{\bf}{\it}
{\bf}{\it}
{\bf}{\rm}

\newcommand{\argmin}{\operatornamewithlimits{argmin}}

\begin{document}

% \begin{frontmatter}
\title{
Error Bounds for Compressed Sensing Algorithms\\
With Group Sparsity: A Unified Approach
}

\author{M.\ Eren Ahsen and M.\ Vidyasagar
\thanks{
MEA is with IBM Research, Thomas J. Watson Research Center, 1101 Route 134 Kitchawan Rd, Yorktown Heights, NY 10598.
MV is with the Systems Engineering Department,
% Erik Jonsson School of Engineering and Computer Science,
University of Texas at Dallas, Richardson, TX 75080.
Emails: mahsen@us.ibm.com, m.vidyasagar@utdallas.edu.
This research was Supported by the National Science Foundation under
Award \#1306630, and
the Cecil H.\ \& Ida Green Endowment at UT Dallas}
}

\maketitle

\begin{abstract}

In compressed sensing, in order to recover a sparse or nearly
sparse vector from possibly noisy measurements, the most popular
approach is $\ell_1$-norm minimization.
Upper bounds for the $\ell_2$- norm of the error between the
true and estimated vectors are given in \cite{Candes08} and
reviewed in \cite{DDEK12}, while bounds for the $\ell_1$-norm
are given in \cite{FR13}.
When the unknown vector is not conventionally sparse but is ``group
sparse'' instead, a variety of alternatives to the $\ell_1$-norm
have been proposed in the literature, including the group LASSO,
sparse group LASSO, and group LASSO with tree structured overlapping groups.
However, no error bounds are available for any of these modified
objective functions.
%Another recent direction is sorted $\ell_1$-norm minimization
%for conventionally sparse vectors.
In the present paper, a unified approach is presented for deriving
upper bounds on the error between the true vector and its approximation,
based on the notion of decomposable and $\g$-decomposable norms.
The bounds presented cover all of the norms mentioned above, and
also provide a guideline for choosing norms in future to accommodate
alternate forms of sparsity.

\end{abstract}

% \end{frontmatter}

\section{Introduction}\label{sec:intro}

The field of ``compressed sensing'' has become very popular in recent years,
with an explosion in the number of papers.
In the interests of brevity, we refer the reader to two recent
% survey paper in
% a volume that collects several papers \cite{DDEK12} and another
papers \cite{DDEK12,NRWY12}, each of which contains an extensive bibliography.
Stated briefly, the core problem in compressed sensing is to approximate a
high-dimensional sparse (or nearly sparse) vector $x$ from a small number
of linear measurements of $x$.
Though this problem has a very long history (see the discussion in
\cite{DDEK12} for example), perhaps it is fair to say that much of
the recent excitement has arisen from \cite{Candes-Tao05},
in which it is shown that if $x$ has no more than $k$ nonzero components,
then by choosing the matrix $A$ to satisfy a condition known as the
restricted isometry property (RIP), it is possible to recover
$x$ exactly by minimizing $\nm z \nm_1$ subject to the constraint that
$Az = y = Ax$.
In other words, under suitable conditions,
among all the preimages of $y = Ax$ under $A$, the
preimage that has minimum $\ell_1$-norm is the sparse signal $x$ itself.
The same point is also made in \cite{Donoho06b}.
Subsequently the RIP was replaced by the null space property
\cite{Cohen-Dahmen-Devore09}, which is actually necessary and sufficient
for the above statement to be true; see \cite[Chapter 4]{FR13}
for precise statements.
In case $y = Ax + \eta$ where $\eta$ is a measurement error and $x$
is either sparse or nearly sparse, one can attempt to recover $x$ by setting
\be\label{eq:11}
\xh := \argmin_{z \in \R^n} \nm z \nm_1 \st \nmeu { y - Az } \leq \e .
\ee
% minimizing the $\ell_1$-norm of $z$
% subject to a constraint on the least squares error $\nmeusq { y - Az }$.
This algorithm is very closely related to 
the LASSO algorithm introduced in \cite{Tibshirani-Lasso}.
Specifically, the only difference between LASSO as in \cite{Tibshirani-Lasso}
and the problem stated above is that the roles of the objective function
and the constraint are reversed.
It is shown (see \cite[Theorem 1.2]{Candes08}) that, under suitable
conditions, the residual error $\nmeu { \xh - x }$ satisfies
an estimate of the form
\be\label{eq:12}
\nmeu { \xh - x } \leq \frac{ C_0 }{\sqrt{k}} \s_k(x, \nmm{\cdot}_1) + C_2 \e ,
\ee
where $\s_k(x, \nmm{\cdot}_1)$ is the ``sparsity index'' of $x$ (defined below),
and $C_0,C_2$ are constants that depend only on the matrix $A$
but not $x$ or $\eta$.
The above bound includes exact signal recovery with noiseless measurements
as a special case,
and is referred to in \cite{Candes08} as ``noisy recovery.''
Along similar lines, it is shown in \cite{FR13} that
\be\label{eq:12a}
\nmm { \xh - x }_1 \leq C_0 \s_k(x, \nmm{\cdot}_1) + C_2 \sqrt{k}  \e ,
\ee
where $C_0$ and $C_2$ are the same as in \eqref{eq:12}.
See the equation just above \cite[Equation (4.16)]{FR13}.

In the world of optimization, the LASSO algorithm has been generalized
in several directions, by modifying the $\ell_1$-norm penalty of
LASSO to some other norm that is supposed to induce a prespecified
sparsity structure on the solution.
Among the most popular sparsity-inducing penalty norms are
the group LASSO \cite{Yuan-Lin-Group-Lasso,Huang-Zhang10},
referred to hereafter as GL,
and the sparse group LASSO \cite{FHT10,SFHT12}, referred to hereafter as SGL.
Now there are versions of these algorithms that permit the groups
to have ``tree-structured'' overlap \cite{Jenetton-et-al11,OJV-Over-GL11}.
%A recent contribution is to replace the usual $\ell_1$-norm
%by the ``sorted'' $\ell_1$-norm \cite{Candes-sorted13}.
%A similar idea is earlier proposed in \cite{Daubechies-et-al10},
%but in the context of the sorted $\ell_2$-norm.

It is therefore natural to ask whether
inequalities analogous to (\ref{eq:12}) and \eqref{eq:12a}
hold when the $\ell_1$-norm
in (\ref{eq:11}) is replaced by other sparsity-inducing norms such
as those mentioned in the previous paragraph.
To the best of the authors' knowledge, no such error bounds are
available in the literature for anything other than $\ell_1$-norm
minimization.
In principle, it is possible to mimic the arguments in \cite{Candes08}
to derive error bounds for each of these algorithms.
However, it would be highly desirable to have a unified theory
of what properties a norm needs to satisfy, in order
that inequalities of the form (\ref{eq:12}) hold.
That is the focus of the present paper.
We present a very general
result to the effect that {\it any\/} compressed sensing algorithm
satisfies error bounds of the form \eqref{eq:12} and \eqref{eq:12a}
provided three conditions are satisfied:
\ben
\item A ``compressibility condition'' holds, which
in the case of $\ell_1$-norm minimization is that
the restricted isometry property (RIP) holds with a sufficiently
small constant.
\item The approximation
norm used to compute the sparsity index of the unknown vector
$x$ is ``decomposable'' as defined subsequently.
\item The penalty norm used to induce the sparsity of the solution, that is,
the norm that is minimized, is ``$\g$-decomposable'' as defined
subsequently.
\een
It will follow as a consequence of this general result that
%sorted $\ell_1$-norm minimization, 
GL, and SGL (without  or with
tree-structured overlapping groups) all
satisfy error bounds of the form \eqref{eq:12}.
In addition to the generality of the results established,
the method of proof is more direct than that in \cite{Candes08,DDEK12}.
In the case of conventional sparsity and $\ell_1$-norm minimization,
the results presented here contain those in \cite{Candes08,DDEK12}
as special cases, and also include a bound on $\nmm{\xh - x}_1$, in
addition the bound on $\nmeu{\xh-x}$.

\section{Preliminaries}\label{sec:prelim}

If $x \in \R^n$, and $\L$ is a subset
of $\N = \{ 1 , \ldots , n \}$, the symbol $x_\L \in \R^n$ denotes
the vector such that $(x_\L)_i = x_i$ if $i \in \L$, and $(x_\L)_i = 0$
if $i \not \in \L$.
In other words, $\xl$ is obtained from $x$ by replacing $x_i$ by zero
whenever $i \not \in \L$.
Also, as is customary, for a vector $u \in \R^n$, its support set
is defined by
\bd
\supp(u) := \{ i : u_i \neq 0 \} .
\ed

Let $k$ be some integer that is fixed throughout the paper.
Next we introduce the notion of a group $k$-sparse set.
Some care is required in doing so, as the discussion following the
definition shows.

\begin{definition}\label{def:GKS}
Let $\G = \{ G_1 , \ldots , G_g\}$ be a partition of
$\N = \{ 1 , \ldots , n \}$, such that $| G_i | \leq k$ for all $i$.
If $S \seq \{ 1 , \ldots , g \}$, define $G_S := \cup_{i \in S} G_i$.
A subset $\L \seq \N$ is said to be \textbf{group $k$-sparse} if
there exists a subset $S \seq \{ 1 , \ldots , g \}$ such that
$\L = G_S$, and in addition, $| \L | \leq k$.
The collection of all group $k$-sparse subsets of $\N$ is denoted by $\GkS$.
A vector $u \in \R^n$ is said to be \textbf{group $k$-sparse}
if its support set $\supp(u)$ is contained in a group $k$-sparse set.
\end{definition}

At this point the reader might ask why a set $\L$ cannot be defined to
be group $k$-sparse if it is a {\it subset\/} of some $G_S$,
as opposed to being {\it exactly equal\/} to some $G_S$.
The reason is that, if every subset of $G_S$ is also called
``group $k$-sparse,''
then in effect \textit{all sets of cardinality $k$} or less can be called
group $k$-sparse, thus defeating the very purpose of the definition.
To see this, let $\L = \{ x_{i_1} , \ldots , x_{i_l} \}$, where $l \leq k$,
so that $| \L | = l \leq k$.
Then, since the sets $G_1 , \ldots , G_g$ partition the index set $\N$,
for each $j$ there exists a set $G_j$ such that $x_{i_j} \in G_j$.
Let $S \seq \{ 1 , \ldots , g \}$ denote the set consisting of all these
indices $j$.
Then $\L \seq G_S$.
So with this modified definition, there would be no difference between
group $k$-sparsity and conventional sparsity.
This is the reason for adopting the above definition.
On the other hand, it is easy to see that if $g = n$ and each set $G_i$
equals the singleton set $\{ i\}$, then group $k$-sparsity reduces to
conventional $k$-sparsity.
Note also that a vector is defined to be group $k$-sparse if its
support {\it is contained in}, though not necessarily equal to,
a group $k$-sparse subset of $\N$.

Suppose $\nmm { \cdot } : \R^n \ap \R_+$ is some norm.
We introduce a couple of notions of decomposability that build upon
an earlier definition from \cite{NRWY12}.

\begin{definition}\label{def:decomp}
The norm $\nmm { \cdot }$ is said to be {\bf decomposable}
with respect to the partition $\G$ if, whenever $u, v \in \R^n$
with $\supp(u) \seq G_{S_u}$, $\supp(v) \seq G_{S_v}$,
and $S_u, S_v$ are disjoint subsets of $\{ 1 , \ldots , g \}$,
it is true that
\be\label{eq:21a}
\nmm { u + v } = \nmm { u } + \nmm { v } .
\ee
\end{definition}

As pointed out in \cite{NRWY12}, because $\nmm { \cdot }$ is a norm,
the relationship (\ref{eq:21a}) {\it always\/} holds with $\leq$ replacing
the equality.
Therefore the essence of decomposability is that the bound is tight
when the two summands are vectors with their support
sets contained in disjoint
sets of the form $G_{S_u},G_{S_v}$.
Note that it is not required for (\ref{eq:21a}) to hold for every
pair of vectors with disjoint supports, only vectors whose support sets
are contained in disjoint unions of group $k$-sparse subsets of $\N$.
For instance, if $\L$ is a group $k$-sparse set, and $u,v$
have disjoint support sets $\supp(u),\supp(v)$ that are both subsets of
$\L$, then there is no requirement that (\ref{eq:21a}) hold.
It is easy to see that the $\ell_1$-norm is decomposable,
and it is shown below that 
the group LASSO and the sparse group LASSO norm are also decomposable. 
%However, the sorted $\ell_1$-norm 
%introduced in \cite{Candes-sorted13} is not decomposable.
To generalize our analysis, we introduce a more general definition of decomposability.

\begin{definition}\label{def:gdecomp}
The norm $\nmm { \cdot }$ is {\bf $\g$-decomposable} with respect to the
partition $\G$ if there exists $\g \in (0,1]$ such that,
whenever $u, v \in \R^n$
with $\supp(u) \seq G_{S_u}$, $\supp(v) \seq G_{S_v}$,
and $S_u, S_v$ are disjoint subsets of $\{ 1 , \ldots , g \}$,
it is true that
\be\label{eq:21}
\nmm { u + v } \geq \nmm { u } + \g \nmm { v } .
\ee
\end{definition}

Note that if the norm $\nmm{\cdot}$ is $\g$-decomposable with
$\g=1$, then (\ref{eq:21}) and the triangle inequality imply that
\bd
\nmm { u + v } \geq \nmm { u } + \nmm { v } \implies \nmm { u + v }
= \nmm { u } + \nmm { v } .
\ed
Therefore decomposability is the same as $\g$-decomposability with $\g = 1$.

Clearly, if $\nmm { \cdot }$ is a decomposable norm, 
then (\ref{eq:21a}) can be applied recursively to show that
if $\L_0 , \L_1 , \ldots , \L_s$ are pairwise disjoint group $k$-sparse sets,
% the corresponding sets $S_0 , \ldots , S_s$ are pairwise disjoint,
and $\supp(u_i) \seq \L_i$, then
\be\label{eq:22}
\left\nm \sum_{i=0}^s u_i \right\nm = \sum_{i=0}^s \nmm { u_i } .
\ee
However, such an equality does not hold 
for $\g$-decomposable functions unless $\g=1$,
which makes the norm decomposable.
On the other hand, by repeated application of (\ref{eq:21})
and noting that $\g \leq 1$, we arrive at the following relationship:
if $\L_0 , \L_1 , \ldots , \L_s$ are pairwise disjoint group $k$-sparse sets,
% the corresponding sets $S_0 , \ldots , S_s$ are pairwise disjoint,
and $\supp(u_i) \seq \L_i$, then
\be\label{eq:22b}
\left\nm \sum_{i=0}^s u_i \right\nm \geq \nmm { u_{\L_0} } 
+ \g \left\nm \sum_{i=1}^s u_i \right\nm .
\ee

Equation (\ref{eq:22}) is somewhat more general than the definition of
decomposability given in \cite{NRWY12},
in that we permit the partitioning of the index set $\N$ into more than
two subsets.
However, this is a rather minor generalization.\footnote{There is
a little bit of flexibility in \cite{NRWY12}
in that one can take two orthogonal subspaces that are not exactly
orthogonal complements of each other; but we will not belabor this point.}

It is now shown that the notions of decomposability and $\g$-decomposability
are general enough to encompass several algorithms such as
group LASSO, sparse group LASSO, either without overlapping groups
or with groups that overlap but have a tree structure.
%, and 
%sorted $\ell_1$-norm minimization.

\begin{lemma}\label{lemma:21}
Let $\G = \{ G_1 , \ldots , G_g \}$ be a partition of
the index set $\N = \{ 1 , \ldots , n \}$.
Let $\nm \cdot \nm_i : \R^{| G_i |} \ap \R_+$ be
\textit{any} norm, and define the corresponding norm on $\R^n$ by
\be\label{eq:63b}
\nmA { x } = \sum_{i=1}^g \nm x_{G_i} \nm_i .
\ee
Then the above norm is decomposable.
\end{lemma}

The proof is omitted as it is obvious.
The key point to note is that the
exact nature of the individual norms $\nm \cdot \nm_i$ is
entirely irrelevant.

By defining the individual norms $\nm \cdot \nm_i$ appropriately,
it is possible to recover the group LASSO \cite{Yuan-Lin-Group-Lasso,
Huang-Zhang10}, the sparse group LASSO \cite{FHT10,SFHT12},
and the overlapping sparse group LASSO with tree-structured norms
\cite{Jenetton-et-al11,OJV-Over-GL11}.

\begin{corollary}\label{corr:21}
The group LASSO norm defined by
\be\label{eq:61}
\nm z \nm_{{\rm GL}} := \sum_{i=1}^g \nmeu { z_{G_i} } .
\ee
is decomposable.
\end{corollary}

\textbf{Proof:}
This corresponds to the choice $\nm \cdot \nm_i = \nmeu { \cdot }$.
Note that some authors use $\nmeu { z_{G_i} } / \sqrt{ | G_i | }$
instead of just $\nmeu { z_{G_i} }$.
This variant is also decomposable, as is easy to see.
\halmos

\begin{corollary}\label{corr:22}
The sparse group LASSO norm defined by
\be\label{eq:62}
\nm z \nm_{{\rm SGL},\mu} := \sum_{i=1}^g
[ (1 - \mu) \nm z_{G_i} \nm_1 + \mu \nmeu { z_{G_i} } ] .
\ee
is decomposable.
\end{corollary}

\textbf{Proof:}
This corresponds to the choice
\bd
\nm z_{G_i} \nm_i = (1 - \mu) \nm z_{G_i} \nm_1 + \mu \nmeu { z_{G_i} } .
\ed
Therefore the norm is decomposable.
\halmos

Next let us turn our attention to the case of ``overlapping'' groups
with tree structure,
as defined in \cite{Jenetton-et-al11,OJV-Over-GL11}.

\begin{corollary}\label{corr:23}
Suppose there are sets $\N_1 , \ldots , \N_l$, each of which
is a subset of $\N$, that satisfy the condition
\be\label{eq:63a}
\N_i \cap \N_j \neq \es \implies ( \N_i \seq \N_j \mbox{ or }
\N_j \seq \N_i ) .
\ee
Define
\bd
\nmA{z} = \sum_{i=1}^l \nmm{ z_{\N_i} }_i ,
\ed
where $\nmm{\cdot}_i : \R^{| \N_i |}  \ap \R_+$ is arbitrary.
Then this norm is decomposable.
\end{corollary}

\textbf{Proof:}
Though it is possible for some of these sets $\N_i$ to overlap,
the condition \eqref{eq:63a} implies that the collection of sets $\N_1 , \ldots , \N_l$
can be renumbered with double indices as $\S_{ij}$,
and arranged in chains of the form
\bd
\S_{11} \seq \ldots \seq \S_{1n_1} , \ldots , 
\S_{s1} \seq \ldots \seq \S_{sn_s} ,
\ed
where the ``maximal'' sets $\S_{in_i}$ must also satisfy (\ref{eq:63a}).
Therefore, given two maximal sets $\S_{in_i}, \S_{jn_j}$,
either they must be the same or they must be disjoint, because it is not
possible for one of them to be a subset of the other.
This shows that the maximal sets $\S_{in_i}$ are pairwise disjoint once
the duplicates are removed,
and together span the total feature set $\N = \{ 1 , \ldots , n \}$.
Thus, in a collection of tree-structured sets, the highest level sets
do not overlap!
Let $g$ denote the number of distinct maximal sets, and define
$G_i = \S_{in_i}$ for $i = 1 , \ldots, g$.
Then $\{ G_1 , \ldots , G_g \}$ is a partition of $\N$,
and each $\N_j$ is a subset of some $G_i$.
Now define a norm $\nmm{\cdot}_{G_j}, j = 1 , \ldots , g$, by
\bd
\nmm{z_{G_j}}_{G_j} = \sum_{\N_i \seq G_j} \nmm{z_{\N_i}}_i .
\ed
Because each $\N_j$ can be a subset of only one $G_i$, it
follows that the above norm is the same as $\nmA{\cdot}$.
Therefore this norm is of the form \eqref{eq:63b} and is thus decomposable.
\halmos

Thus to summarize, the group LASSO norm, the sparse group LASSO norm, and
the penalty norms defined in 
\cite{Jenetton-et-al11,OJV-Over-GL11} are all decomposable.

With this preparation we can define the sparsity indices and optimal
decompositions.
Given an integer $k$, let $\GkS$ denote the collection of all group
$k$-sparse subsets of $\N = \{ 1 , \ldots , n \}$, and define
\be\label{eq:23}
\s_{k,\G} (x , \nmm { \cdot } ) := \min_{ \L \in \GkS} \nmm { x - \xl }
= \min_{ \L \in \GkS} \nmm { \xloc } 
\ee
to be the {\bf group $k$-sparsity index} of the vector $x$ with respect
to the norm $\nmm { \cdot }$ and the group structure $\G$.
Since the collection of sets $\GkS$ is finite (though it could be huge),
we are justified in writing $\min$ instead of $\inf$.
Once we have the definition of the sparsity index, it is natural to define
the next notion.
Given $x \in \R^n$, and a norm $\nmm { \cdot }$,
we call $\{ x_{\L_0} , x_{\L_1} , \ldots , x_{\L_s} \}$
an {\bf optimal group $k$-sparse decomposition} of $x$ if
$\L_i \in \GkS$ for $i = 0 , \ldots , s$, and in addition
\bd
\nmm { \xloc } = \min_{\L \in \GkS} \nmm { x - x_\L } ,
\ed
\bd
\nmm { x_{\L_i^c} } = \min_{\L \in \GkS} \left\nm x - \sum_{j=0}^{i-1} x_{\L_j} 
- x_\L \right\nm , i = 1 , \ldots , s .
\ed

There are some wrinkles in group sparsity that do not have any analogs
in conventional sparsity.
Specifically, suppose $x \in \R^n$ and that $\{ x_{\L_0} , x_{\L_1} ,
\ldots , x_{\L_s} \}$ is an optimal $k$-sparse ({\it not\/} optimal
{\it group\/} $k$-sparse) decomposition of $x$
with respect to $\nm \cdot \nm_1$.
Then $x_{\L_0}$ consists of the $k$ largest components of $x$ by
magnitude, $x_{\L_1}$ consists of the next $k$ largest, and so on.
One consequence of this is that
\bd
\min_j | (x_{\L_i})_j | \geq \max_j | (x_{\L_{i+1}})_j | , \fa i .
\ed
Therefore
\be\label{eq:58}
\nmeu { x_{\L_{i+1} } } \leq \sqrt{k} \nm x_{\L_{i+1} } \nm_\infty
\leq \frac{1}{ \sqrt{k} } \nm x_{\L_i} \nm_1 .
\ee
This is the equation just above \cite[Equation(10)]{Candes08}.
However, when we take optimal {\it group\/} $k$-sparse decompositions, this
inequality is no longer valid.
For example, suppose $\nmP { \cdot } = \nm \cdot \nm_1$, 
let $n = 4 , g = 2 , k = 2$ and
\bd
G_1 = \{ 1 , 2 \} , G_2 = \{ 3, 4 \} ,
x = [ \ba{cccc} 1 & 0.1 & 0.6 & 0.6 \ea ]^t .
\ed
Then it is easy to verify that $s = 2$, and
\bd
\L_0 = \{ 3, 4 \} = G_2 , \L_1 = \{ 1 , 2 \} = G_ 1 ,
\ed
\bd
x_{\L_0} = [ \ba{cccc} 0 & 0 & 0.6 & 0.6 \ea ]^t ,
% \ed
% \bd
x_{\L_1} = [ \ba{cccc} 1 & 0.1 & 0 & 0 \ea ]^t .
\ed
Here we see that the largest element of $x_{\L_1}$ is in fact larger
than the smallest element of $x_{\L_0}$.
However, we do not have the freedom to ``swap'' these elements
as they belong to different sets $G_i$.
A more elaborate example is the following:
Let $n = 8, g = 4 , k = 4$, and
\bd
x = [ \ba{cccccccc} 0.1 & 1 & 0.2 & 0.3 & 0.4 & 0.5 & 0.4 & 0.7 \ea ] ,
\ed
\bd
G_1 = \{ 1 \} , G_2 = \{ 2, 3, 4 \} , G_3 = \{ 5, 6 \} , G_4 = \{ 7, 8 \} .
\ed
Then 
\bd
\L_0 = G_3 \cup G_4 , \L_1 = G_1 \cup G_2 .
\ed
Note that $x_{G_2}$ has higher $\ell_1$-norm than any other $x_{G_j}$.
However, since $G_2$ has cardinality $3$, it can only be paired with $G_1$,
and not with $G_3$ or $G_4$, in order that the cardinality of the union
remain less than $k = 4$.
And $\nm x_{G_1 \cup G_2} \nm_1 < \nm x_{G_3 \cup G_4} \nm_1$.
Therefore an optimal group $k$-sparse decomposition of $x$ is
$x_{G_3 \cup G_4}$ followed by $x_{G_1 \cup G_2}$.

\section{Problem Formulation}\label{sec:prob}

The general formulation of a compressed sensing algorithm
makes use of three distinct norms:
\bit
\item
$\nmA{ \cdot }$ is the \textit{approximation norm}
that is used to measure the quality of the approximation.
Thus, for a vector $x \in \R^n$, the quantity $\s_{k,\G} (x , \nmA { \cdot } )$
is the sparsity index used throughout.
It is assumed that $\nmA{ \cdot }$ is a \textit{decomposable norm}.
\item
$\nmP { \cdot }$ is the \textit{penalty norm}
that is minimized to induce a desired sparsity structure on the solution.
It is assumed that $\nmP { \cdot }$
is \textit{$\g$-decomposable for some $\g \in (0,1]$}. 
\item
$\nmeu { \cdot }$, which is the standard Euclidean or $\ell_2$-norm, and
is used to constrain the measurement matrix via the group restricted
isometry property (GRIP).
\eit

The prototypical problem formulation is this:
Suppose $x \in \R^n$ is an unknown vector,
$A \in \R^{m \times n}$ is a measurement matrix, 
$y = Ax + \eta$ is a possibly noise-corrupted measurement vector in $\R^m$,
and $\eta \in \R^m$ is the measurement error.
It is presumed that $\nmeu { \eta } \leq \e $, where $\e$ is a known
prior bound.
To estimate $x$ from $y$, we solve the following optimization problem:
\be\label{eq:31}
\xh = \argmin_{z \in \R^n} \nmP { z } \st \nmeu { y - Az } \leq \e .
\ee
The penalty norm $\nmP { \cdot }$ that is
minimized in order to determine an approximation to $x$ need not
be the same as the approximation norm $\nmA { \cdot }$
used to compute the group $k$-sparsity index.
If $\nmP{\cdot}$ is the $\ell_1$-, group LASSO, or sparse group LASSO
norm, then we take $\nmA{\cdot} = \nmP{\cdot}$.
The objective is to determine error bounds of the form\footnote{The symbol $A$ is unfortunately doing
double duty, representing the approximation norm as well as the measurement
matrix.
After contemplating various options, it was decided to stick to this
notation, in the hope that the context would make clear which usage is meant.}
\be\label{eq:32}
\nmeu{ \xh - x } \leq D_1 \s_{k,\G} ( x , \nmA { \cdot } ) + D_2 \e ,
\ee
or of the form
\be\label{eq:32a}
\nmA{ \xh - x } \leq D_3 \s_{k,\G} ( x , \nmA { \cdot } ) + D_4 \e ,
\ee
for some appropriate constants $D_1$ through $D_4$.

The interpretation of the inequality (\ref{eq:32}) in this general setting 
is the same as in \cite{Candes08,DDEK12}.
Suppose the vector $x$ is group $k$-sparse, so that 
$\s_{k,\G} ( x , \nmA { \cdot } ) = 0$.
Then an ``oracle'' that knows the actual support set of $x$
can approximate $x$ through computing a generalized inverse of the columns
of $A$ corresponding to the support of $x$, and the resulting residual
error will be bounded by a multiple of $\e$.
Now suppose the algorithm satisfies (\ref{eq:32}).
Then (\ref{eq:32}) implies that the residual error
achieved by the algorithm is bounded by a universal constant times that 
achieved by an oracle.
Proceeding further, (\ref{eq:32}) also implies that if measurements
are noise-free so that $\e = 0$, then the estimate $\xh$ equals $x$.
In other words, the algorithm achieves exact recovery of group $k$-sparse
vectors under noise-free measurements.
Similar remarks apply to the interpretation of the bound \eqref{eq:32a}.

Throughout the paper, we shall be making use of four constants:
\be\label{eq:33}
a := \min_{\L \in \GkS} \min_{x_\L \neq 0} \frac{ \nmP { x} }
{ \nmA {x } } ,
b := \max_{\L \in \GkS} \max_{x_\L \neq 0} \frac{ \nmP { \xl } }
{ \nmA {\xl } } ,
\ee
\be\label{eq:34}
c := \min_{\L \in \GkS} \min_{x_\L \neq 0} \frac{ \nmA { \xl } }
{ \nmeu {\xl } } ,
d := \max_{\L \in \GkS} \max_{x_\L \neq 0} \frac{ \nmA { \xl } }
{ \nmeu {\xl } } .
\ee
Note that these constants depend on the sparsity structure being
used.
For instance, in conventional sparsity, as shown below, $a = b = c = 1$
and $d = \sqrt{k}$.
The factor $\sqrt{k}$ is ubiquitous in conventional sparsity, and
as shown below, this is where it comes from.

Suppose for instance that $\nmA { \cdot } = \nmP { \cdot } = \nm \cdot \nm_1$,
which is the approximation as well as penalty norm used in LASSO.
Therefore $a = b = 1$.
Since $|\L| \leq k$ for all $\L \in \GkS$, we have by Schwarz's inequality that
\bd
\nm v \nm_1 \leq \sqrt{k} \nmeu { v }
\ed
whenever $\supp(v) \seq \L \in \GkS$.
In the other direction, we can write
\bd
v = \sum_{i \in \supp(v)} v_i \eb_i ,
\ed
where $\eb_i$ is the $i$-th unit vector.
Therefore by the triangle inequality
\bd
\nmeu { v } \leq \sum_{i \in \supp(v)} \nmeu { v_i \eb_i }
\leq \sum_{i \in \supp(v)} | v_i | = \nm v \nm_1 ,
\ed
and these bounds are tight.
Therefore
\bd
1 = c \leq d = \sqrt{k} .
\ed
Estimates of these constants for other 
sparsity-inducing
norms are given in Section \ref{sec:proofs}.

\section{Main Results}\label{sec:main}

In this section we present the main results of the paper.
The following definition of the restricted isometry property (RIP)
is introduced in \cite{Candes-Tao05}.

\begin{definition}\label{def:RIP}
Suppose $A \in \R^{m \times n}$.
Then we say that $A$ satisfies the {\bf Restricted Isometry Property (RIP)}
of order $k$ with constant $\d_k$ if
\be\label{eq:41a}
(1 - \d_k) \nm u \nm_2^2 \leq \langle u , Au \rangle \leq
(1 + \d_k) \nm u \nm_2^2 , \fa u \in \SI_k ,
\ee
where $\SI_k$ denotes the set of all $u \in \R^n$ such that
$| \supp(u) | \leq k$.
\end{definition}

The first step is to extend the notion of the restricted isometry property
(RIP) to the group restricted isometry property (GRIP).

\begin{definition}\label{def:GRIP}
A matrix $A \in \R^{m \times n}$ is said to satisfy the \textbf{group
restricted isometry property (GRIP)}
of order $k$ with constant $\d_k \in (0,1)$ if
\be\label{eq:51}
1 - \d_k \leq \min_{ \L \in \GkS} \min_{\supp(z) \seq \L} 
\frac{ \nm Az \nm_2^2 }{ \nm z \nm_2^2} \leq
\max_{ \L \in \GkS} \max_{\supp(z) \seq \L}
\frac{ \nm Az \nm_2^2 }{ \nm z \nm_2^2} \leq 1 + \d_k .
\ee
\end{definition}

Definition \ref{def:GRIP} shows that the group RIP constant $\d_k$
can be smaller than the standard RIP constant
in Definition \ref{def:RIP}, because the various maxima and minima
are taken over only group $k$-sparse sets, and not all 
subsets of $\N$ of cardinality $k$.
Probabilistic methods for constructing a measurement matrix 
$A \in \R^{m \times n}$ that satisfies GRIP with specified order
$k$ and constant $\d$ are discussed in Section \ref{sec:sample}.
It is shown that GRIP can be achieved with a smaller value of $m$ than RIP.

In order to state the main results, we introduce a technical lemma.

\begin{lemma}\label{lemma:51}
Suppose $h \in \R^n$, that $\L_0 \in GkS$ is arbitrary, and let
$h_{\L_1} , \ldots , h_{\L_s}$ be an optimal group $k$-sparse decomposition
of $\hloc$ with respect to the decomposable approximation norm $\nmA{ \cdot }$.
Then there exists a constant $f$ such that
\be\label{eq:59}
\sum_{j=2}^s \nmeu { h_{\L_j} } \leq \frac{1}{f} \nmA { \hloc } .
\ee
\end{lemma}

{\bf Proof:}
It is already shown in \cite[Equation (11)]{Candes08}, 
\cite[Lemma A.4]{DDEK12} that
\bd
\sum_{j=2}^s \nmeu{h_{\L_j}} \leq \frac{1}{\sqrt{k}} \nmm{\hloc}_1 .
\ed
Therefore, in the case of conventional sparsity,
where $\nmA{\cdot} = \nmP{\cdot} = \nmm{\cdot}_1$, one can take $f = \sqrt{k}$.
In the case of group sparsity, it follows from the definition of the
constant $c$ in \eqref{eq:34} that
\bd
\nmeu{ h_{\L_j}} \leq \frac{1}{c} \nmA{ h_{\L_j} } , j = 2 , \ldots , s .
\ed
Therefore
\bd
\sum_{j=2}^s \nmeu { h_{\L_j} } \leq \frac{1}{c}
\sum_{j=2}^s \nmA { h_{\L_j} } 
\leq \sum_{j=1}^s \nmA { h_{\L_j} } = \frac{1}{c} \nmA { \hloc } ,
\ed
where the last step follows from the decomposability of $\nmA{\cdot}$.
\halmos

Now we state the main theorem for the general optimization problem
as stated in \eqref{eq:31}, and several corollaries for conventional
sparsity, group LASSO, sparse group LASSO
%, and sorted $\ell_1$-norm
minimization.
All of these are stated at once, followed by a general discussion.

\begin{theorem}\label{thm:52}
Suppose that
\ben
\item The norm $\nmA{\cdot}$ is decomposable.
\item The norm $\nmP{\cdot}$ is $\g$-decomposable for some $\g \in (0,1]$.
\item The matrix $A$ satisfies GRIP of order $2k$ with constant $\d_{2k}$.
\item
Suppose the ``compressibility condition''
\be\label{eq:511}
% \frac{ \sqrt{2} \d_{2k} }{ 1 - \d_{2k} } < \frac { fa \g }{bd} ,
\d_{2k} < \frac{ f a \g }{ \sqrt{2} + f a \g / bd }
\ee
holds, where $d$ is defined in \eqref{eq:34} and $f$ is defined in
Lemma \ref{lemma:51}.
\een
Define
\be\label{eq:512}
\xh = \argmin_{z \in \R^n} \nmP { z } \st \nmeu { y - Az } \leq \e .
\ee
Then
\be\label{eq:513}
\nmeu{ \xh - x } \leq D_1 \s_{k,\G} ( x , \nmA{ \cdot}) + D_2 \e ,
\ee
where
\be\label{eq:514}
D_1 = \frac{ r(1+\g)}{f} \cdot \frac{ 1 + (\sqrt{2}-1 ) \d_{2k} }
{1 - ( 1 + \sqrt{2} rd/f) \d_{2k} } ,
\ee
\be\label{eq:515}
D_2 = 2 ( 1 + rd/f) \frac{ \sqrt{ 1 + \d_{2k} } }
{1 - ( 1 + \sqrt{2} rd/f) \d_{2k} } .
\ee
Further,
\be\label{eq:516}
\nmA{ \xh - x } \leq D_3 \s_{k,\G} ( x , \nmA{ \cdot}) + D_4 \e ,
\ee
where
\be\label{eq:517}
D_3 =  r(1+\g) \cdot \frac{ 1 + (\sqrt{2} d/f -1 ) \d_{2k} }
{1 - ( 1 + \sqrt{2} rd/f) \d_{2k} } ,
\ee
\be\label{eq:518}
D_4 = 2 ( 1 + rd) \frac{ \sqrt{ 1 + \d_{2k} } }
{1 - ( 1 + \sqrt{2} rd/f) \d_{2k} } .
\ee
\end{theorem}

\begin{corollary}\label{corr:51}
\textbf{(Conventional Sparsity)}
Define
\be\label{eq:81}
\xh_{{\rm CS}} = \argmin_z \nmm{z}_1 \st \nmeu{ y - Az } \leq \e .
\ee
Then Theorem \ref{thm:52} applies with $\nmA{\cdot} = \nmP{\cdot} =
\nmm{\cdot}_1$,
\be\label{eq:82}
a = 1 , b = 1 ,  c = 1 , d = \sqrt{k} , f = \sqrt{k} , \g = 1 .
\ee
Therefore the compressibility condition \eqref{eq:511} becomes
\be\label{eq:82aa}
\d_{2k} < \sqrt{2} - 1 .
\ee
This leads to the error bounds
\be\label{eq:83}
\nmeu{\xh_{{\rm CS}} - x} \leq D_2 \s_k( x, \nmm{\cdot}_1 )
+ D_2 \e ,
\ee
\be\label{eq:84}
\nmm{\xh_{{\rm CS}} - x}_1 \leq D_3 \s_k( x, \nmm{\cdot}_1 )
+ D_4 \e ,
\ee
where
\bd
D_1 = \frac{2}{\sqrt{k}} 
\frac{ 1 + ( \sqrt{2} - 1 ) \d_{2k} } { 1 - (1 + \sqrt{2} \d_{2k} ) } ,
D_2 = 4 \frac{ 1 + \sqrt{\d_{2k} } } { 1 - (1 + \sqrt{2} \d_{2k} ) } ,
\ed
\bd
D_3 = 2 \frac{ 1 + ( \sqrt{2} - 1 ) \d_{2k} } { 1 - (1 + \sqrt{2} \d_{2k} ) }
\s_k( x, \nmm{\cdot}_1 ) ,
D_4 = 4 \sqrt{k} \frac{ 1 + \sqrt{\d_{2k} } } { 1 - (1 + \sqrt{2} \d_{2k} ) } .
\ed
\end{corollary}

\begin{corollary}\label{corr:52}
\textbf{(Group LASSO)}
Suppose $\{ G_1 , \ldots , G_g \}$ is a partition of $\N = \{ 1 , \ldots , 
n \}$, and that $l_{{\rm min}} \leq | G_j | \leq k$ for all $j$.
Let $s_{{\rm max}} = \lfloor k/l_{{\min}} \rfloor$, and define
the group LASSO norm
\be\label{eq:85}
\nmm{z}_{{\rm GL}} = \sum_{j=1}^g \nmeu{z_{G_j}} .
\ee
Define the estimate
\be\label{eq:86}
\xh_{{\rm GL}} = \argmin_z \nmm{z}_{{\rm GL}} \st \nmeu{y - Az} \leq \e .
\ee
Then Theorem \ref{thm:52} applies with $\nmA{\cdot} = \nmP{\cdot} =
\nmm{\cdot}_{{\rm GL}}$,
\be\label{eq:87}
a = 1 , b = 1 ,  c = 1 , d = \sqrt{s_{{\rm max}}} , f = 1 , \g = 1 .
\ee
Therefore the compressibility condition \eqref{eq:511} becomes
\be\label{eq:82b}
\d_{2k} < \frac{1}{\sqrt{ 2 s_{{\rm max}} } + 1 }
\ee
This leads to the error bounds 
\bd
\nmeu{ \xh_{{\rm GL}} - x} \leq D_1 \s_k( x , \nmm {\cdot}_{{\rm GL}} )
+ D_2 \e ,
\ed
and
\bd
\nmm{ \xh_{{\rm GL}} - x}_{{\rm GL}} \leq D_3 \s_k( x , \nmm {\cdot}_{{\rm GL}} )
+ D_4 \e ,
\ed
where
\bd
D_1 = 2 \frac{ 1 + ( \sqrt{2} - 1 ) \d_{2k} } { 1 - (1 + \sqrt{2 s_{{\rm max}}} ) \d_{2k}  } ,
D_2 = 4 \frac{ 1 + \sqrt{\d_{2k} } } { 1 - (1 + \sqrt{2 s_{{\rm max}}}  )\d_{2k}  } ,
\ed
\bd
D_3 = 2 \frac{ 1 + ( \sqrt{2} - 1 ) \d_{2k} } { 1 - (1 + \sqrt{2 s_{{\rm max}}}  )\d_{2k}  } ,
D_4 = 4 \sqrt{s_{{\rm max}} } \frac{ 1 + \sqrt{\d_{2k} } } { 1 - (1 + \sqrt{2 s_{{\rm max}}}  )\d_{2k}  } .
\ed
\end{corollary}

\begin{corollary}\label{corr:53}
\textbf{(Sparse Group LASSO)}
Suppose $\{ G_1 , \ldots , G_g \}$ is a partition of $\N = \{ 1 , \ldots , 
n \}$, and that $l_{{\rm min}} \leq | G_j | \leq l_{{\rm max}}$ for all $j$.
Let $s_{{\rm max}} = \lfloor k/l_{{\min}} \rfloor$, and define
the sparse group LASSO norm
\be\label{eq:88}
\nmm{z}_{{\rm SGL},\mu} = \sum_{j=1}^g
[ (1 - \mu) \nmm{z_{G_j}}_1  + \mu \nmeu{z_{G_j}} ] .
\ee
Define the estimate
\be\label{eq:89}
\xh_{{\rm SGL}} = \argmin_z \nmm{z}_{{\rm SGL},\mu} \st \nmeu{y - Az} \leq \e .
\ee
Then Theorem \ref{thm:52} applies with $\nmA{\cdot} = \nmP{\cdot} =
\nmm{\cdot}_{{\rm SGL}}$,
\be\label{eq:90}
a = 1 , b = 1 ,  c = 1 ,
d = ( 1 - \mu) \sqrt{l_{{\rm max}} } + \mu \sqrt{s_{{\rm max}}} ,
f = 1 , \g = 1 .
\ee
Therefore the compressibility condition \eqref{eq:511} becomes
\be\label{eq:82a}
\d_{2k} < \frac{ d }{ \sqrt{2} + d } ,
\ee
where $d$ is defined in \eqref{eq:90}.
This leads to the error bounds
\bd
\nmeu{ \xh_{{\rm SGL}} - x} \leq D_1 \s_k( x , \nmm_{\cdot}_{{\rm SGL}} )
+ D_2 \e ,
\ed
and
\bd
\nmm{ \xh_{{\rm SGL}} - x}_{{\rm SGL}} \leq D_3 \s_k( x , \nmm_{\cdot}_{{\rm GL}} )
+ D_4 \e ,
\ed
where the constants $D_1$ through $D_4$ are the same as in Corollary
\ref{corr:52} with the term $\sqrt{s_{{\rm max}} }$ replaced by
$d$ as shown in \eqref{eq:90}.
\end{corollary}

Before presenting the proofs of these bounds, we briefly discuss their
implications.
\ben
\item In the case of conventional sparsity, the bounds on $\nmeu{\xh - x}$
and $\nmm{\xh-x}_1$
reduce to 
those proved earlier in \cite{Candes08,DDEK12,FR13}.
To the best of the authors' knowledge, there are no bounds of
the form \eqref{eq:12} and \eqref{eq:12a} available for other penalty norms.
Therefore the bounds in Theorem \ref{thm:52} contain known bounds
as special cases and some new bounds as well.
\item In the case of conventional sparsity, the upper bound on
$\nmm{\xh-x}_1$ is precisely $\sqrt{k}$ times the upper bound on
$\nmeu{\xh-x}$.
Note that if the vector $\xh-x$ is $k$-sparse, then by Schwarz' inequality
it would follow that $\nmm{\xh-x}_1 \leq \sqrt{k} \nmeu{\xh-x}$.
It is therefore interesting that a similar relationship holds even though
the residual error $\xh-x$ need not be $k$-sparse.
\item In the case of the group LASSO norm, the key parameter is 
$s_{{\rm max}}$, the largest
number of sets $G_i$ that can comprise any group $k$-sparse set.
If each set $G_i$ is a singleton, then $s_{{\rm max}} = k$.
\item The only difference between the bounds for the group LASSO and the
sparse group LASSO norms is in the parameter $d$.
\een

\section{Proofs of Main Results}\label{sec:proofs}

The proof of Theorem \ref{thm:52} depends on a few preliminary lemmas.

Lemma \ref{lemma:52} should be compared with 
\cite[Lemma 2.1]{Candes08}, \cite[Lemma A.3]{DDEK12}.

\begin{lemma}\label{lemma:52}
Suppose $A \in \R^{m \times n}$
satisfies the group RIP of order $2k$ with constant
$\d_{2k}$, and that $u,v$ are group $k$-sparse
with supports contained in disjoint group $k$-sparse subsets of $\N$.
Then
\be\label{eq:510}
| \langle Au , Av \rangle | \leq \d_{2k} \nm u \nm_2 \cdot \nm v \nm_2 .
\ee
\end{lemma}

{\bf Proof:}
Since we can divide through by $\nm u \nm_2 \cdot \nm v \nm_2$,
an equivalent statement is the following:
If $u,v$ are group $k$-sparse with supports contained in disjoint
group $k$-sparse subsets of $\N$,
and $\nm u \nm_2 = \nm v \nm_2 = 1$, then
\bd
| \langle Au , Av \rangle | \leq \d_{2k} .
\ed
Now the assumptions guarantee that $u \pm v$ are both group $2k$-sparse.
Moreover $u^t v = 0$ since they have disjoint support.
Therefore $\nm u \pm v \nm_2^2 = 2$.
So the group RIP implies that
\bd
2 (1 - \d_{2k} ) \leq \nm Au \pm Av \nm_2^2 \leq 2 (1 + \d_{2k} ) .
\ed
Now the parallelogram identity implies that
\bd
| \langle Au , Av \rangle | =
\left| \frac{ \nm Au + Av \nm_2^2 - \nm Au - Av \nm_2^2 } {4} \right|
\leq \d_{2k} .
\ed
This is the desired conclusion.
\halmos

\begin{lemma}\label{lemma:54}
Suppose $h \in \R^n$, that $\L_0 \in GkS$ is arbitrary, and let
$h_{\L_1} , \ldots , h_{\L_s}$ be an optimal group $k$-sparse decomposition
of $\hloc$ with respect to the approximation norm $\nmA{ \cdot }$.
Define $\L = \L_0 \cup \L_1$.
Then
\be\label{eq:521}
\nmeu { \hl } \leq \frac{ \sqrt{2} \d_{2k} } {f (1 - \d_{2k}) } \nmA { \hloc }
+ \frac { \sqrt{(1 + \d_{2k}) } } { (1 - \d_{2k})  } \nmeu { Ah } .
\ee
\end{lemma}

The proof closely mimics that of
\cite[Lemma 1.3]{DDEK12}.
But it is presented in detail, in the interests of completeness.

{\bf Proof:}
Note that $\hl$ is group $2k$-sparse.
Therefore by the definition of the group RIP property, it follows that
\bd
(1 - \d_{2k}) \nmeusq { \hl } \leq \nmeusq { A \hl }
\leq (1 + \d_{2k}) \nmeusq { \hl } .
\ed
Next, observe that
\bd
\nmeusq { A \hl } = \langle A \hl , A \hl \rangle .
\ed
So we will work on a bound for the right side.
Note that
\bd
\langle A \hl , A \hl \rangle
= \langle A \hl , Ah \rangle - \langle A \hl , A \hlc \rangle .
\ed
Next by (\ref{eq:516}) and Schwarz's inequality, it follows that
\beq
| \langle A \hl , A \hlc \rangle |
& \leq & \left| \sum_{i=0}^1
\sum_{j=2}^s \langle A h_{\L_i} , A h_{\L_j} \rangle \right|
\nonumber \\
& \leq & \d_{2k} [ \nmeu { h_{\L_0} } + \nmeu { h_{\L_1} } ]
\sum_{j=2}^s \nmeu { h_{\L_j} }
\nonumber \\
& \leq & \frac{ \sqrt{2} \d_{2k} } {f} \nmeu { \hl } \nmA { \hlc } . \nonumber
\eeq
In the above, we use the known inequality
\bd
\nmeu { h_{\L_0} } + \nmeu { h_{\L_1} } 
\leq \sqrt{2} \nmeu { h_{\L_0} + h_{\L_1} } 
= \sqrt{2} \nmeu { \hl } ,
\ed
because $h_{\L_0}$ and $h_{\L_1}$ are orthogonal.
Next
\bd
| \langle A \hl , Ah \rangle |
\leq \nmeu { A \hl } \cdot \nmeu { A h }
\leq \sqrt{ (1 + \d_{2k}) } \nmeu { \hlo } \cdot \nmeu { A h } .
\ed
Combining everything gives
\beq
(1 - \d_{2k}) \nmeusq { \hl } & \leq & \nmeusq { A \hl } \nonumber \\
& \leq & | \langle A \hl , Ah \rangle | + | \langle A \hl , A \hlc \rangle |
\nonumber \\
& \leq & \frac{ \sqrt{2} \d_{2k} } {f} \nmeu { \hl } \nmA { \hlc }
+ \sqrt{ (1 + \d_{2k}) } \nmeu { \hl } \cdot \nmeu { A h } \nonumber .
\eeq
Dividing both sides by $(1 - \d_{2k}) \nmeu { \hl }$ leads to (\ref{eq:521}).
\halmos

{\bf Proof of Theorem \ref{thm:52}:}
Define $\xh$ as in \eqref{eq:512}, and define $h = \xh - x$, so that
$\xh = x + h$.
The optimality of $\xh$ implies that $\nmP{x} \geq \nmP{\xh} = \nmP{x + h}$.

Let $\{ x_{\L_0} , x_{\L_1} , \ldots , x_{\L_s} \}$ be an optimal group
$k$-sparse decomposition of $x$.
Then the triangle inequality and the optimality of $\xh$ together imply that
\be\label{eq:41}
\sum_{i=0}^s \nmP{ x_{\L_i} } \geq \nmP { x } \geq \nmP{ x + h } .
\ee
Now the $\g$-decomposability of $\nmP{\cdot}$ implies that
\beq
\nmP { x + h } & \geq & \nmP{ \xlo + \hlo + \xloc + \hloc } \nonumber \\
& \geq & \nmP{ \xlo + \hlo } + \g \sum_{i=1}^s \nmP{ x_{\L_i} + h_{\L_i} } 
\nonumber \\
& \geq & \nmP{\xlo} - \nmP{\hlo}
+ \g \sum_{i=1}^s [ \nmP{ h_{\L_i} } - \nmP{ x_{\L_i} } ] .
\label{eq:42}
\eeq
Combining \eqref{eq:41} and \eqref{eq:42}, cancelling the common term
$\nmP{\xlo}$, and rearranging leads to
\bd
\g \sum_{i=1}^s \nmP{ h_{\L_i} } \leq \nmP{\hlo}
+ ( 1 + \g) \sum_{i=1}^s \nmP{ x_{\L_i} } .
\ed
Next we make use the definition of the constants $a$ and $b$ from
\eqref{eq:33}, the decomposability of $\nmA{\cdot}$, and the triangle
inequality.
This leads to
\beq
a \g \nmA{\hloc} & = & a \g \sum_{i=1}^s \nmA{h_{\L_i} } \nonumber \\
& \leq & \g \sum_{i=1}^s \nmP{ h_{\L_i} } \nonumber \\
& \leq & \nmP{\hlo} + ( 1 + \g) \sum_{i=1}^s \nmP{ x_{\L_i} } \nonumber \\
& \leq & b \nmA{\hlo} + b ( 1 + \g) \sum_{i=1}^s \nmA{ x_{\L_i} } \nonumber \\
& = &  b \nmA{\hlo} + b ( 1 + \g) \nmA{\xloc} \nonumber \\
& = & b \nmA{\hlo} + b ( 1 + \g) \s_A , \nonumber \\
\eeq
where $\s_A$ is shorthand for $\s_{k,\G}(x,\nmA{\cdot})$, the group
$k$-sparsity index of $x$.
Dividing both sides by $a \g$ gives
\be\label{eq:522b}
\nmA{\hloc} \leq r \nmA{\hlo} + r ( 1 + \g) \s_A ,
\ee
where $r = b/a\g$.
Next, it follows from the definition of $d$ in \eqref{eq:34} that
\bd
\nmA{\hlo} \leq d \nmeu{\hlo} \leq d \nmeu{\hl} ,
\ed
where as before $\L = \L_0 \cup \L_1$.
Substituting into the previous bound gives
\be\label{eq:522}
\nmA{\hloc} \leq r d \nmeu{\hl} + r ( 1 + \g) \s_A .
\ee
This is the first of two inequalities that we require.

Next, both $x$ and $\xh$ are feasible for the optimization problem
in \eqref{eq:512}.
This implies that
% the feasibility of both $x$ and $\xh$ implies that
\bd
\nmeu { Ah } \leq \nmeu{A(\xh - x)} \leq \nmeu{A \xh - y}
+ \nmeu{Ax - y} \leq 2 \e .
\ed
Therefore (\ref{eq:521}) now becomes
\be\label{eq:522a}
\nmeu { \hl } \leq \frac{ \sqrt{2} \d_{2k} } {f (1 - \d_{2k}) } \nmA { \hloc }
+ \frac { 2 \sqrt{ 1 + \d_{2k} } } { (1 - \d_{2k})  } \e .
\ee
Define the symbols
\be\label{eq:522c}
g = \frac{ \sqrt{2} \d_{2k} } { (1 - \d_{2k}) } ,
r_2 = \frac { 2 \sqrt{ 1 + \d_{2k} } } { (1 - \d_{2k})  } ,
\ee
so that $g$ and $r_2$ depend only the GRIP constant $\d_{2k}$.
% and observe that $\nmA{\hloc} = \s_A$.
Therefore \eqref{eq:523} can be expressed compactly as
\be\label{eq:523}
\nmeu{\hl} \leq (g/f) \nmA { \hloc } + r_2 \e .
\ee
This is the second inequality we require.

The inequalities (\ref{eq:522}) and (\ref{eq:523}) can be
written as a vector inequality, namely
\bd
\left[ \ba{cc} 1 & -r d \\ - g/f & 1 \ea \right]
\left[ \ba{l} \nmA { \hloc } \\ \nmeu { \hl } \ea \right]
\leq \left[ \ba{c} r (1 + \g)  \\ 0 \ea \right] \s_A
+ \left[ \ba{c} 0 \\ r_2 \ea \right] \e .
\ed
The coefficient matrix on the left side has a strictly positive inverse
if its determinant $1 - grd/f$ is positive.
So the ``compressibility condition'' is $g < f/rd$, which is the same
as \eqref{eq:511}.
Moreover, if $1 - grd/f > 0$,
then one can infer from the above vector inequality that
\beq
\left[ \ba{l} \nmA { \hloc } \\ \nmeu { \hl } \ea \right]
& \leq & \frac{1}{ 1 - grd/f } \left[ \ba{cc} 1 & r d \\ g/f & 1 \ea \right]
\left\{ \left[ \ba{c} r (1+\g) \\ 0 \ea \right] \s_A
+ \left[ \ba{c} 0 \\ r_2 \ea \right] \e \right\} \nonumber \\
& = & \frac{1}{ 1 - grd/f } \left\{ \left[ \ba{c} 1 \\ g/f \ea \right] 
r (1+\g) \s_A
+ \left[ \ba{c} r d \\ 1 \ea \right] r_2 \e \right\} . \nonumber
\eeq
Now by (\ref{eq:59}),
\bd
\nmeu { \hlc } \leq \sum_{j=2}^s \nmeu { h_{\L_j} }
\leq \frac{1}{f} \nmA { \hloc } .
\ed
Therefore, since $h = \hl + \hlc$, the triangle inequality implies that
\beq
\nmeu { h } 
& \leq & \nmeu { \hlc } + \nmeu { \hl } \nonumber \\
& \leq & \frac{1}{f} \nmA { \hloc } + \nmeu { \hl } \nonumber \\
& \leq & \frac{1}{ 1 - grd/f } [ \ba{cc} 1/f & 1 \ea ]
\left\{ \left[ \ba{c} 1 \\ g/f \ea \right] r (1+\g) \s_A
+ \left[ \ba{c}r d \\ 1 \ea \right] r_2 \e \right\}  \nonumber \\
& = & \frac{1}{ 1 - grd/f } [ r (1+\g) ( 1/f + g/f) \s_A + (1 + (r d)/f) r_2 \e ] . \nonumber
\eeq
Substituting for the various constants and clearing leads to the
bound in \eqref{eq:513}.

To derive the bound \eqref{eq:516} on $\nmA{\xh-x}$, we adopt
the same strategy of deriving a vector inequality and then inverting
the coefficient matrix.
We already have from \eqref{eq:522b} that
\bd
\nmA{\hloc} \leq r \nmA{\hlo} + r ( 1 + \g) \s_A .
\ed
Next, it follows from the definition of $d$ in \eqref{eq:34} and
\eqref{eq:522a} that
\bd
\nmA{\hlo} \leq d \nmeu{\hlo} \leq d \nmeu{\hl} 
\leq \frac{gd}{f} \nmA{\hloc} + r_2 \e , 
\ed
where $g$ and $r_2$ are defined in \eqref{eq:522c}.
These two inequalities can be combined into the vector inequality
\bd
\left[ \ba{cc} 1 & -r \\ -gd/f & 1 \ea \right]
\left[ \ba{c} \nmA{\hloc} \\ \nmA{\hlo} \ea \right] \leq
\left[ \ba{l} r(1+\g) \s_A \\ r_2 \e \ea \right] .
\ed
Though the coefficient matrix is different, the determinant is still
$1 - rdg/f$.
Therefore, if \eqref{eq:511} holds, then the coefficient matrix has 
a positive inverse.
In this case we can conclude that
\beq
\nmA{h} & \leq & [ \ba{cc} 1 & 1 \ea ]
\left[ \ba{c} \nmA{\hloc} \\ \nmA{\hlo} \ea \right] \nonumber \\
& \leq & \frac{1}{1 - rdg/f} [ \ba{cc} 1 & 1 \ea ]
\left[ \ba{cc} 1 & r \\ gd/f & 1 \ea \right]
\left[ \ba{l} r(1+\g) \s_A \\ r_2 \e \ea \right] . \nonumber
\eeq
After clearing terms, this is the bound \eqref{eq:516}.
\halmos

\textbf{Proof of Corollary \ref{corr:51}:}
If both $\nmA{\cdot}$ and $\nmP{\cdot}$ are equal, it is obvious that
$a = b = 1$, as defined in \eqref{eq:33}.
Next, it is a ready consequence of Schwarz' inequality that $c = 1$
and $d = \sqrt{k}$, as defined in \eqref{eq:34}.
Next, it is shown in \cite[Equation (11)]{Candes08},
\cite[Lemma A.4]{DDEK12} that $f$ defined in \eqref{eq:59} equals $\sqrt{k}$.
Because $\nmP{\cdot}$ is decomposable, we can take $\g = 1$.
Substituting these values into the bound \eqref{eq:513} through
\eqref{eq:516} establishes
the desired bounds \eqref{eq:83} and \eqref{eq:84}.
\halmos

\textbf{Proof of Corollary \ref{corr:52}:}
Let $\nmA{\cdot} = \nmP { \cdot } = \nm \cdot  \nm_{{\rm GL}}$.
Then since $\nmP{\cdot}=\nmA{\cdot}$, we have $a = b = 1$.
To calculate $c$ and $d$, define $l_{{\rm min}}$ to be the smallest
cardinality of any $G_i$,
and define $s_{{\rm max}} := \lfloor k/ l_{{\rm min}} \rfloor$.
Now suppose that $\L \in \GkS$.
Specifically, suppose $\L = G_{i_1} \cup \ldots \cup G_{i_s}$.
Then clearly
\bd
\nm z_\L \nm_{{\rm GL}} = \sum_{j=1}^s \nmeu { z_{G_{i_j}} } ,
\ed
while
\bd
\nmeu { z_\L } = \left( \sum_{j=1}^s \nmeusq { z_{G_{i_j}} } \right)^{1/2} .
\ed
Thus, if we define the $s$-dimensional vector $v \in \R_+^s$ by
\bd
v = [ \nmeu { z_{G_{i_j}} } , j = 1 , \ldots , s ] ,
\ed
then
\bd
\nm z_\L \nm_{{\rm GL}} = \nm v \nm_1 , 
\nmeu { z_\L } = \nmeu { v } .
\ed
Now it is easy to see that
\bd
\nmeu { v } \leq \nm v \nm_1 \leq \sqrt{s} \nmeu { v } .
\ed
Moreover, it is clear that the integer $s$, denoting the number of
distinct sets that make up $\L$, cannot exceed $s_{{\rm max}}$.
This shows that
\be\label{eq:64}
1 \leq c_{{\rm GL}} \leq d_{{\rm GL}} \leq \sqrt{s_{{\rm max}} } .
\ee
As shown in the proof of Lemma \ref{lemma:51}, in the case of
group sparsity, one can only take $f = c = 1$.
Finally, because $\nmP{\cdot}$ is decomposable, it follows that $\g = 1$.
Substituting these values into \eqref{eq:513} through \eqref{eq:518}
leads to the desired bounds.
\halmos

\textbf{Proof of Corollary \ref{corr:53}:}
In this case
$\nmA { \cdot } = \nmP { \cdot } = \nm \cdot \nm_{{\rm SGL},\mu}$.
Because both norms are equal, it follows that $a = b = 1$.
To calculate $c$ and $d$, suppose $\L = G_{i_1} \cup \ldots \cup G_{i_s}$.
Let $l_{{\rm max}}$ denote the largest cardinality of any $G_i$.
Then
\bd
\nmeu { z_{G_{i_j}} } \leq \nm z_{G_{i_j}} \nm_1 \leq
\sqrt{ l_{{\rm max}} } \nmeu { z_{G_{i_j}} } ,
\ed
whence
\be\label{eq:65}
\sum_{j=1}^s \nmeu { z_{G_{i_j}} } \leq \sum_{j=1}^s \nm z_{G_{i_j}} \nm_1 \leq
\sqrt{ l_{{\rm max}} } \sum_{j=1}^s \nmeu { z_{G_{i_j}} } .
\ee
Combining (\ref{eq:64}) and (\ref{eq:65}) leads to
\bd
\nmeu { z_\L } \leq \nm z_\L \nm_{{\rm SGL},\mu} \leq
[ ( 1 - \mu ) \sqrt{ l_{{\rm max}} } + \mu \sqrt{s_{{\rm max}} } ]
\nmeu { z_\L } .
\ed
Therefore
\be\label{eq:66}
1 \leq c_{{\rm SGL},\mu} \leq d_{{\rm SGL},\mu} 
\leq ( 1 - \mu ) \sqrt{ l_{{\rm max}} } + \mu \sqrt{s_{{\rm max}} } .
\ee
Again, in the case of group sparsity one has to take $f = c = 1$.
Finally, because $\nmP{\cdot}$ is decomposable, we can take $\g = 1$.
Substituting these values into \eqref{eq:513} through \eqref{eq:518}
leads to the desired bounds.
\halmos

\section{Bounds on the Number of Measurements}\label{sec:sample}

In this section we study the following problem:
Suppose a matrix $A \in \R^{m \times n}$ is constructed by drawing
$mn$ i.i.d.\ samples of a fixed random variable $X$.
Suppose we are specified integers $n, k \ll n$, and real numbers
$\d , \zeta \in (0,1)$.
The objective is to determine a lower bound on $m$ such that
$A$ satisfies GRIP or order $k$ with constant $\d$, with probability
no smaller than $1 - \zeta$.

The approach here follows \cite{BDDW08,FR13}.
Recall that a zero-man random variable $X$ is said to be \textbf{sub-Gaussian}
if there exist constants $\al,\beta$ such that
\be\label{eq:71}
\Pr \{ | X | > \e \} \leq \al \exp(-\beta \e^2) , \fa \e > 0 .
\ee
A normal random variable satisfies \eqref{eq:71} with $\al = 2, \beta = 0.5$.
Suppose in addition that $X$ has unit variance, and define
$A \in \R^{m \times n}$ by drawing $mn$ i.i.d.\ samples of $X/m$.
Then it is known (\cite[Lemma 9.8]{FR13}) that
\bd
\Pr \{ | \nmeusq { Au } - \nmeusq{u} | > \e \nmeusq{u} 
\leq 2 \exp(-mc \e^2),
\ed
where
\be\label{eq:72}
c = \frac{\beta^2}{4 \al + 2 \beta} .
\ee
With this background, we can begin to address the problem under study.

\begin{lemma}\label{lemma:71}
Given integers $n,k \ll n$ and a real number $\d \in (0,1)$,
and any collection $\J$ of subsets of $\N = \{ 1 , \ldots , n \}$
such that $|T| \leq k \fa t \in \J$.
Let $X$ be a zero-mean, unit variance, sub-Gaussian random variable
satisfying \eqref{eq:71}, and let $A \in \R^{m \times n}$ consist of
$mn$ i.i.d.\ samples of $X$.
Then
\be\label{eq:73}
( 1- \d ) \nmeusq{x_T} \leq \nmeusq {A x_T} \leq (1+\d) \nmeusq{x_T}
\fa T \in \J , \fa x \in \R^n
\ee
with probability no smaller than $1 - \zeta$, where $\zeta$ is given by
\be\label{eq:74}
\zeta = 2 |\J| \left( \frac{12}{\th} \right)^k \exp(-m c \th^2) ,
\ee
where $c$ is defined in \eqref{eq:72} and
\be\label{eq:75}
\th = 1 - \sqrt{1 - \d} .
\ee
\end{lemma}

\textbf{Proof:} It is shown in \cite[Lemma 5.1]{BDDW08} that, for a given
fixed index set $T \seq \N$ with $|T| \leq k$, the inequality
\be\label{eq:76}
( 1- \th ) \nmeu{x_T} \leq \nmeu {A x_T} \leq (1+\th) \nmeu{x_T} ,
\fa x \in \R^n ,
\ee
with probability no smaller than $1 - \zeta'$, where
\be\label{eq:77}
\zeta' = 2 \left( \frac{12}{\th} \right)^k \exp(-m c \th^2) .
\ee
However, the inequality \eqref{eq:76}
does not quite match the definition of RIP or GRIP,
because the inequality involves $\nmeu{Ax_T}$ and not $\nmeusq{Ax_T}$.
Therefore, in order to convert \eqref{eq:77} into \eqref{eq:73},
we need to have
\bd
1 - \d \leq (1 - \th)^2 , \mbox{ and } (1 + \th)^2 \leq 1 + \d ,
\ed
or equivalently,
\bd
\th \leq \max \{ 1 - \sqrt{1 - \d} , \sqrt{1 + \d} - 1 \} .
\ed
It is elementary to show that the first term is always larger than the
second, so that \eqref{eq:76} implies \eqref{eq:73}
provided $\th$ is defined as in \eqref{eq:75}.

Next, suppose the collection $\J$ is specified.
Then \cite[Lemma 5.1]{BDDW08} implies that \eqref{eq:76} holds for each
fixed set with probability no smaller than $1 - \zeta'$.
Therefore the union of events bound shows that \eqref{eq:73} holds with
probability no smaller than $1 - |\J| \zeta'$, where $\zeta'$ is defined
in \eqref{eq:77}.
The proof is completed by noting that $\zeta$ defined in \eqref{eq:74}
is precisely $|\J| \zeta'$.
\halmos

Now we are ready to give estimates for the integer $m$.

\begin{theorem}\label{thm:71}
Suppose integers $n, k \ll n$ are specified, together with real numbers
$\d,\zeta \in (0,1)$.
Let $X$ be a sub-Gaussian zero-mean unit-variance random variable,
and define the constant $c$ as in \eqref{eq:72}.
Let $A \in \R^{m \times n}$ consist of $mn$ i.i.d.\ random samples of $X/m$.
Define $\th$ as in \eqref{eq:75}.
Then
\ben
\item $A$ satisfies RIP of order $k$ with constant $\d$, with probability
no smaller than $1 - \zeta$, provided
\be\label{eq:78}
m_{{\rm S}} \geq \frac{1}{c \th^2} \left[ \log \frac{2}{\zeta}
+ k \left( \log \frac{en}{k} + \log \frac{12}{\th} \right) \right] .
\ee
\item Suppose $\{ G_1 , \ldots , G_g \}$ is a partition of $\N = \{ 1, 
\ldots , n \}$, where $l_{{\rm min}} \leq |G_i| \leq k$ for all $i$.
Define $s_{{\rm max}} = \lfloor k/l_{{\rm min}} \rfloor$.
Then $A$ satisfies GRIP of order $k$ with constant $\d$, with probability
no smaller than $1 - \zeta$, provided
\be\label{eq:78a}
m_{{\rm GS}} \geq \frac{1}{c \th^2} \left[ \log \frac{2}{\zeta}
+ s_{{\rm max}}  \log \frac{eg}{s_{{\rm max}} } + k \log \frac{12}{\th} \right] .
\ee
\een
\end{theorem}

\textbf{Proof:}
Suppose a set $S$ consists of $s$ elements, and that $t < s$.
Then the number of distinct subsets of $S$ with $t$ or fewer elements is
given by
\bd
\sum_{i=0}^t \left( \ba{c} s \\ i \ea \right) \leq \left(
\frac{es}{t} \right)^t ,
\ed
where the bound is a part of Sauer's lemma,
which can be found in many places, out of which
\cite[Theorem 4.1]{MV-03} is just one reference.
To prove (1), note that the number of distinct subsets of $\N$
with $k$ or fewer elements is bounded by $(en/k)^k$ by Sauer's lemma.
Therefore, given $n, k, \d , \zeta$, one can choose $m$ large enough that
\bd
2 \left( \frac{en}{k} \right)^k
 \left( \frac{12}{\th} \right)^k \exp(-m c \th^2) \leq \zeta,
\ed
which is equivalent to \eqref{eq:78}, and $A$ would satisfy RIP
or ofder $k$ with constant $\d$ with probability no less than $1 - \zeta$.
To prove Item 2, note that every group $k$-sparse set is a union of
at most $s_{{\rm max}}$ sets among $G_1 , \ldots , G_g \}$.
Therefore the number of group $k$-sparse subsets of $\N$ is
bounded by $(eg/s_{{\rm max}})^s_{{\rm max}}$.
Therefore, given $n, k, \d , \zeta$, one can choose $m$ large enough that
\bd
2 \left( \frac{eg}{s_{{\rm max}} } \right)^{s_{{\rm max}} }
 \left( \frac{12}{\th} \right)^k \exp(-m c \th^2) \leq \zeta,
\ed
which is equivalent to \eqref{eq:78a}, and $A$ would satisfy GRIP
or ofder $k$ with constant $\d$ with probability no less than $1 - \zeta$.
\halmos

One of the nice features of these bounds \eqref{eq:77} and \eqref{eq:78}
is that in both cases the confidence level $\zeta$
enters through the logarithm, so that
$m$ increases very slowly as we decrease $\zeta$.
This is consistent with the well-known maxim in statistical learning
theory that ``confidence is cheaper than accuracy.''

% As to whether group sparsity offers any advantage over pure sparsity,
% the evidence is not conclusive.
Next we compare the number of measurements required with conventional
versus group sparsity.
It is pointed out in \cite{Huang-Zhang10} that if random projections
are used to construct $A$, then satisfying the group RIP requires
fewer samples than satisfying RIP.
In particular, suppose all groups have the same size $s$, implying that
$n = gs$ where $g$ is the number of groups.
Suppose also that $k$ is a multiple of $s$, say $k = sr$.
Then satisfying the group RIP condition requires only $O(k + r \log g)$
random projections, whereas satisfying the RIP requires $O(k \log n)$
random projections.
The bounds in Theorem \ref{thm:71} generalize these observations,
as they do not require that all groups must be of the
same size, or that either $n$ or $k$ be a multiple of the group size.
Note that, when $\d$ is very small, $\th \approx \d/2$.
Therefore a comparison of (\ref{eq:77}) and (\ref{eq:78}) shows that
$m_{{\rm S}}$ is $O( k \log n)/\d^2$, whereas 
$m_{{\rm GS}}$ is $O( k + s_{{\rm max}} \log g)/\d^2$.
This is the generalization of the
the term involving $s_{{\rm max}}$ will dominate the term involving $k$.
So in principle group sparsity would require fewer measurements than
conventional sparsity.
However, since $s_{{\rm max}}$ is multiplied by $\log g$,
$g$ would have to be truly enormous in order for group sparsity to lead
to substantially smaller values for $m$ than conventional sparsity.

The important point is that, unless is $n$ is extremely large,
neither of the bounds \eqref{eq:77} or \eqref{eq:78} leads to 
a value of $m$ that is smaller than $n$.
To illustrate this last comment, let us apply the bounds 
from Theorem \ref{thm:71} to typical numbers from
microarray experiments in cancer biology.
Accordingly, we take $n = 20,000$, which is roughly equal to the
number of genes in the human body and the number of measured quantities
in a typical experiment, and we take $k = 20$, which is a typical
number of key biomarkers that we hope will explain most of the observations.
Since $\d \leq \sqrt{2} - 1$ is the compressibility condition for
conventional sparsity, we take $\d = 1/4 = 0.25$.
We partition the set of $20,000$ measurements into $g = 6,000$
sets representing the number of pathways that we wish to study,
and we take $l_{{\rm min}} = 4$, meaning that the shortest pathway
of interest has four genes.
Therefore we can take $s_{{\rm max}} = \lfloor k/ l_{{\rm min}} \rfloor = 5$.
Finally, let us take $\zeta = 10^{-8}$.
With these numbers, it is readily verified that
\bd
m_{{\rm S}} = 53,585 ,
m_{{\rm GS}} = 29,978 .
\ed
In other words, both values of $m$ are {\it larger than\/} $n$!
Therefore one can only conclude that these bounds for $m$
are too coarse to be of practical use at least in computational biology,
though perhaps they might be of use in other applications where $n$ is
a few orders of magnitude larger.
Interestingly, the ``deterministic'' approach to the construction
of $A$ presented in \cite{Devore07} leads to smaller values of $m$,
though in theory $m$ increases as a fractional power of $n$
as opposed to $\log n$.
However, the method in \cite{Devore07} does not offer any advantage
for group sparsity over conventional sparsity.

\section{Conclusions}\label{sec:concl}

In this paper we have presented a unified approach for deriving
upper bounds between the true but unknown sparse (or nearly sparse)
vector and its approximation, when the vector is recovered by
minimizing a norm as the objective function.
The unified approach presented here contains the previously known
results for $\ell_1$-norm minimization as a special case, and is also
sufficiently general to encompass most of the norms that are currently
proposed in the literature, including %the sorted $\ell_1$-norm,
group LASSO norm, sparse group LASSO norm, and the group LASSO norm
with tree-structured overlapping groups.
Estimates for the number of measurements required are derived
for group sparse vectors, and are shown to be smaller than for
conventionally sparse vectors, when the measurement matrix is
constructed using a probabilistic approach.

\section*{Acknowledgement}

The authors thank Mr.\ Shashank Ranjan for his careful reading of an
earlier version of the papers.

\bibliographystyle{elsarticle-num}

\bibliography{Comp-Sens}

\end{document}